\documentclass[conference]{IEEEtran}
\IEEEoverridecommandlockouts
\usepackage{cite}
\usepackage{amsmath,amssymb,amsfonts}
\usepackage{algorithmic}
\usepackage{graphicx}
\usepackage{textcomp}
\usepackage{xcolor}
\def\BibTeX{{\rm B\kern-.05em{\sc i\kern-.025em b}\kern-.08em
    T\kern-.1667em\lower.7ex\hbox{E}\kern-.125emX}}
\begin{document}

\title{Efficient Interdependent Systems Recovery Modeling with DeepONets*\\
{\footnotesize \textsuperscript{*}Note: This is a work in progress paper.}\\
\thanks{This research is supported through the INL Laboratory Directed Research \& Development (LDRD) Program under DOE Idaho Operations Office Contract DE-AC07-05ID14517.}
}

\author{\IEEEauthorblockN{Somayajulu L. N. Dhulipala}
\IEEEauthorblockA{\textit{Computational Mechanics and Materials} \\
\textit{Idaho National Laboratory}\\
Idaho Falls, ID USA \\
Som.Dhulipala@inl.gov\\
ORCID: 0000-0002-0801-4250}
\and
\IEEEauthorblockN{Ryan C. Hruska}
\IEEEauthorblockA{\textit{Infrastructure Analysis} \\
\textit{Idaho National Laboratory}\\
Idaho Falls, ID USA \\
Ryan.Hruska@inl.gov\\
ORCID: 0000-0003-4141-0308}
}

\maketitle

\begin{abstract}
Modeling the recovery of interdependent critical infrastructure is a key component of quantifying and optimizing societal resilience to disruptive events. However, simulating the recovery of large-scale interdependent systems under random disruptive events is computationally expensive. Therefore, we propose the application of Deep Operator Networks (DeepONets) in this paper to accelerate the recovery modeling of interdependent systems. DeepONets are ML architectures which identify mathematical operators from data. The form of governing equations DeepONets identify and the governing equation of interdependent systems recovery model are similar. Therefore, we hypothesize that DeepONets can efficiently model the interdependent systems recovery with little training data. We applied DeepONets to a simple case of four interdependent systems with sixteen states. DeepONets, overall, performed satisfactorily in predicting the recovery of these interdependent systems for out of training sample data when compared to reference results.
\end{abstract}


\section{Introduction}

Infrastructure systems are the backbone of modern societies and are critical for well-functioning communities. Driven by economies of scale these infrastructure systems and their supporting supply chains are often automated and optimized to allow for continued growth at minimum cost. Hazards such as the COVID-19 Pandemic, Hurricane Maria, and recent cyber-attacks have exposed that this practice can quickly result in overtaxed infrastructure systems, that if disrupted lead to widespread goods and services shortages that transverse local, state, and national borders. Further, these infrastructure systems have co-evolved into highly interconnected systems-of-systems which the 1997 Presidential Commission’s Report on Critical Infrastructure Protection (PCCIP) highlighted were becoming increasingly vulnerable to cascading, escalating, and common cause failures \cite{PCCIP1997a}. Since the report was published, significant research has been conducted to gain a better understanding of infrastructure resilience which has enhanced our ability to mitigate vulnerabilities and recover systems following natural or man-made events. However, events like the 2021 Texas Polar Vortex, where a complex series of cascading disruptions to power, natural gas, and water infrastructure demonstrate that our ability to routinely identify and mitigate cross-sector vulnerabilities from an all hazards perspective is critical but remains an open and difficult problem.

This is primarily because critical infrastructures display a wide range of spatial, temporal, operational, organizational, and interdependent characteristics which can affect their ability to adapt to changing conditions. The inherent complexity of these systems can introduce subtle interactions and feedback mechanisms that often lead to unintended behavior and consequences during disruption \cite{Carey2021a}. Central to understanding the resilience of SoS is the ability to simulate SoS behavior and calculate recovery curves under uncertain future disruptive events. These recovery curve simulations further allow for an inference of quantitative resilience metrics at both the systems and SoS levels.

While the simulation of recovery curves is a precursor to quantifying resilience, simulating the recovery of a SoS with numerous interdependent systems under uncertain disruptive events is computationally expensive. To this end, machine learning (ML) is proving to be a valuable tool in accelerating the analysis of computationally expensive tasks. In general, ML models can be classified as data-driven, physics-based, or hybrid depending on their reliance on data and the embedding of physics into their architectures. Deep Operator Networks (DeepONets) were proposed by Lu et al. \cite{Lu2021a} recently to identify mathematical operators from data. Across several computational problems, DeepONets predict the dynamics of the underlying systems with good accuracy while being computationally efficient.

The novelty of this paper lies in the use of DeepONets for modeling the recovery of SoS. We specifically use DeepONets and not other ML architectures because the form of governing equations DeepONets learn and the governing equation for modeling SoS recovery are both similar. Therefore, we hypothesize that DeepONets can accurately predict SoS recovery with little training data, thereby, accelerating the recovery and resilience assessment of interdependent critical infrastructure. While DeepONets have been previously only applied to predict the solutions of differential equations, they have not been applied to infrastructure resilience problems like this study demonstrates.

Sections \ref{sec:sos} and \ref{sec:deeponet} present a mathematical model for SoS recovery and review DeepONets, respectively. Section \ref{sec:results} presents the results on applying DeepONets to predict the recovery of a sixteen state SoS. Finally, section \ref{sec:conc} discusses the conclusions and future work. 

\section{Mathematical model for system-of-systems recovery}\label{sec:sos}

A SoS is a composition of multiple interacting systems. For example a municipal water system that provides water to a power generating station, that powers a substation, that powers the water treatment plant form a SoS. Recovery modeling plans an important role in evaluating the resilience of a SoS, thus many recovery models such as deterministic and semi-deterministic \cite{Cimellaro2010a} have been proposed. A stochastic recovery model for SoS is attractive in that it considers the full range of uncertainties in the individual systems recoveries and their interdependencies and propagates these to the SoS recovery. As such, the SoS recovery characterization is probabilistic and is a function of time. At any time $t$, the stochastic recovery model predicts the probability that the SoS is full functional. Burton et al. \cite{Burton2016a}, Dhulipala et al. \cite{Dhulipala2020a}, and Dhulipala et al. \cite{Dhulipala2021a} are a few studies which have proposed and applied the stochastic recovery model to predict the recovery of residential and commercial infrastructures subjected to hazards like earthquakes and hurricane winds. A brief overview is provided herein.

\begin{figure}[htbp]
\centerline{\includegraphics[scale=0.4]{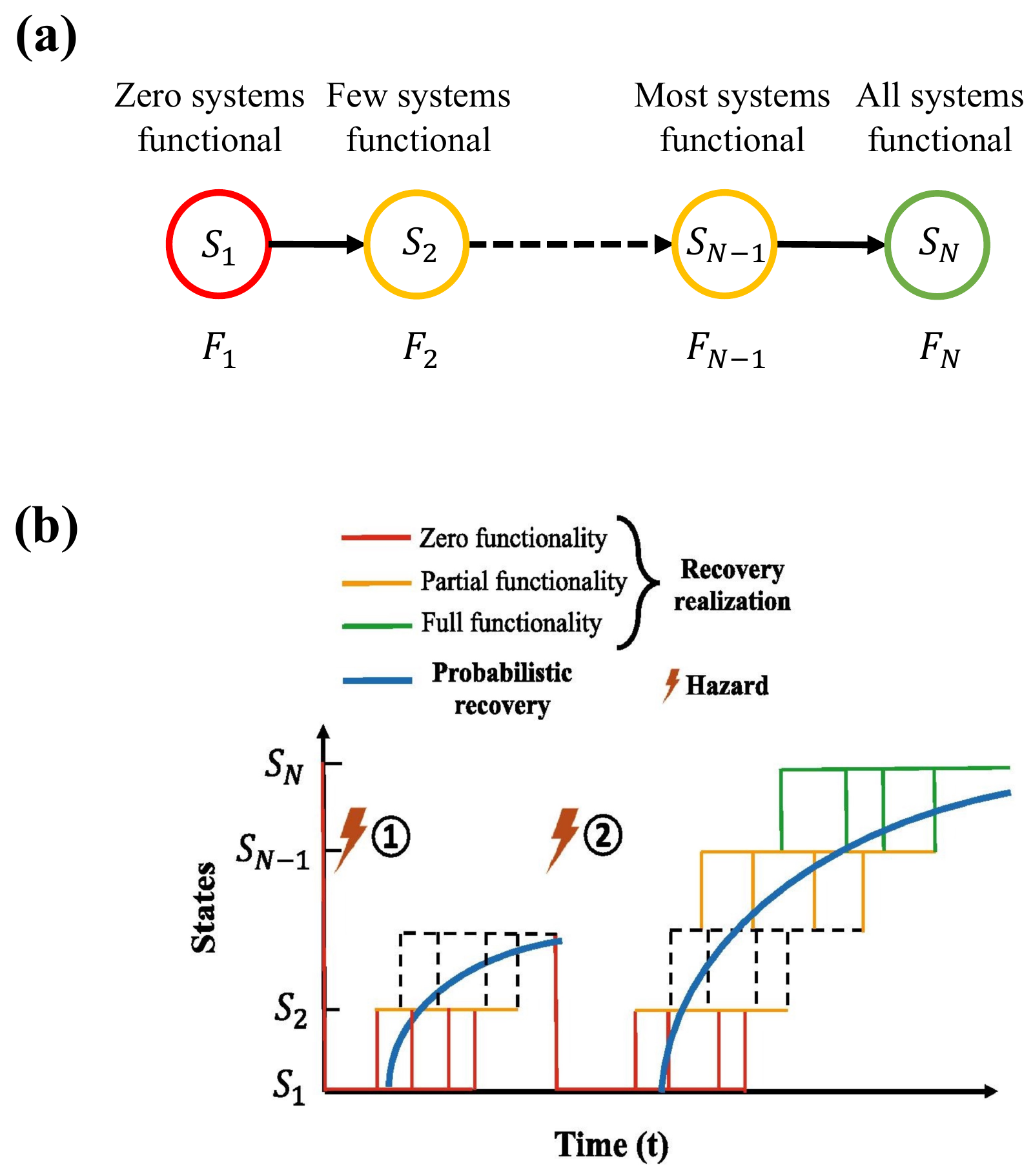}}
\caption{(a) Description of the SoS functionality recovery as a sequence of transitions between discrete functionality states at random times. An increasing state index represents more systems in the SoS are functional. (b) Discrete realizations of the SoS recovery and the SoS probabilistic recovery curve which considers the ensembles of all discrete realizations. This figure is adapted from Dhulipala et al. \cite{Dhulipala2021a}.}
\label{fig:SoS_Recovery}
\end{figure}

Stochastic recovery of SoS can be characterized based on how quickly the individual systems recover after a disruption (or hazard). As such, as presented in Figure \ref{fig:SoS_Recovery}(a), the functionality of the SoS can be described through a discrete set of states. The lowest state $S_1$ is one when none of the individual systems are functional and the SoS functionality $F_1 = 0$. As the state index increases $\{S_2, \dots, S_{N-1},~S_N\}$, more and more systems become functional and the SoS functionality increases (i.e., $F_1 < F_2 < \dots < F_{N-1} < F_N$). The highest state $S_N$ is associated with the highest SoS functionality; typically $F_N = 1$. As stated earlier, in a stochastic recovery model, each individual system transitions from being not functional to full functional at random times. Therefore, as presented in Figure \ref{fig:SoS_Recovery}(b), while a single realization of the systems transitioning to full functionality is discrete, the probabilistic recovery of the SoS considers the ensembles of all discrete realizations and is continuous with time.    

The conceptual model of SoS recovery in Figure \ref{fig:SoS_Recovery} can be described as a discrete state continuous time Markov process. The goal is to quantify a transition probability matrix $\pmb{R}(t)$ an element of which $(R_{ij})$ at any time $t$ represents the probability of transitioning to state $j$ $S_j$ from state $i$ $S_i$. The transition probability matrix can be obtained by solving the Markov-renewal equation \cite{Dhulipala2020a}:

\begin{equation}
    \label{eqn:SoS_Rec_1}
    \begin{aligned}
    &\pmb{R}(t) = \pmb{W}(t) + \int_0^t \pmb{\Phi}^\prime(\tau)~\pmb{R}(t-\tau)~d\tau, \\ 
    &{W}_{ij}(t) = \delta_{ij}\Big(1-\sum_{j=1}^N~\Phi_{ij}(t)\Big)\\
    \end{aligned}
\end{equation}

\noindent where, $\pmb{\Phi}(t)$ is the kernel matrix composed of the individual systems recovery curves as a function of time and $\delta_{ij}$ is the Kronecker delta function. Once the transition probability matrix is obtained, the SoS functionality recovery can be defined as:

\begin{equation}
    \label{eqn:SoS_Rec_2}
    \mathcal{F}(t) = \pmb{I}^\top~\pmb{R}(t)~\pmb{F}
\end{equation}

\noindent where, $\pmb{I}$ is the initial state vector and $\pmb{F}$ is the functionality vector. The vector $\pmb{I}$ contains the probabilities that the SoS is in one of states $\{S_1,~S_2,\dots,~S_N\}$ right after the disruption. It is considered herein that the SoS starts to always recover from $S_1$; that is, $\pmb{I} = \{1,~0,\dots,0\}$. The vector $\pmb{F}$ contains the functionality values of the different states $\{S_1,~S_2,\dots,~S_N\}$. For interdependent systems, the vector $\pmb{F}$ can be modeled by a Bayesian network \cite{Dhulipala2021a}. The components of $\pmb{F}$ denote the joint probability that the SoS is functional and the systems are either functional or not depending upon the recovery state definition. Substituting equation \eqref{eqn:SoS_Rec_1} in \eqref{eqn:SoS_Rec_2}, SoS recovery can be expressed as:

\begin{equation}
    \label{eqn:SoS_Rec_3}
    \mathcal{F}(t) = \pmb{I}^\top~\pmb{W}(t)~\pmb{F} + \int_0^t \pmb{I}^\top~\pmb{\Phi}^\prime(\tau)~\pmb{R}(t-\tau)~\pmb{F}~d\tau
\end{equation}

Modeling SoS recovery through equation \eqref{eqn:SoS_Rec_3} requires the solution of the transition probability matrix from the Markov-renewal equation in \eqref{eqn:SoS_Rec_1}. The Markov-renewal equation is a Volterra nonlinear integral equation. One standard method to solve is to use Laplace and inverse Laplace transform. But this method can be intractable for a large number of SoS states, representing several interacting systems. Dhulipala et al. \cite{Dhulipala2021a} proposed a Monte Carlo method to solve equation \eqref{eqn:SoS_Rec_1}. While Monte Carlo provides a generic framework for handling SoS with an arbitrary number of states, it is computationally expensive. To alleviate this computational burden, we explore the use of DeepONets \cite{Lu2021a} which are reviewed subsequently. 

\section{Review of DeepONets}\label{sec:deeponet}

As described in Lu et al. \cite{Lu2021a}, DeepONets learn operators from datasets. Let $G$ be an operator taking an input function $u$ and $G(u)$ is the output function. At any input $y$, $G(u)(y)$ is a real number. DeepONets can be used to identify nonlinear dynamic systems with outputs of the form \cite{Lu2021a}:

\begin{equation}
    \label{eqn:DONet_1}
    G(u)(y) = \pmb{s}_0 + \int_a^t \pmb{g}\big((Gu)(x),~u(x),~x\big)~dx
\end{equation}

\noindent where, $\pmb{s}_0$ is the initial condition at $a$ and $\pmb{g}\big(.,~.,~.\big)$ can be a convolution operator. Equation \eqref{eqn:SoS_Rec_3} and equation \eqref{eqn:DONet_1} have a similar form in that $\mathcal{F}(t) = G(u)(y)$, $\pmb{I}^\top~\pmb{W}(t)~\pmb{F} = \pmb{s}_0$, $\pmb{g}\big((Gu)(x),~u(x),~x\big) = \pmb{I}^\top~\pmb{\Phi}^\prime(\tau)~\pmb{R}(t-\tau)~\pmb{F}$, and $a = 0$. In equation \eqref{eqn:SoS_Rec_3}, the inputs are the individual systems recovery functions $\phi(t)$ arranged in the matrix $\pmb{\Phi}(t)$ and the time $t$. The output is the SoS recovery at $t$.

\begin{figure}[htbp]
\centerline{\includegraphics[scale=0.32]{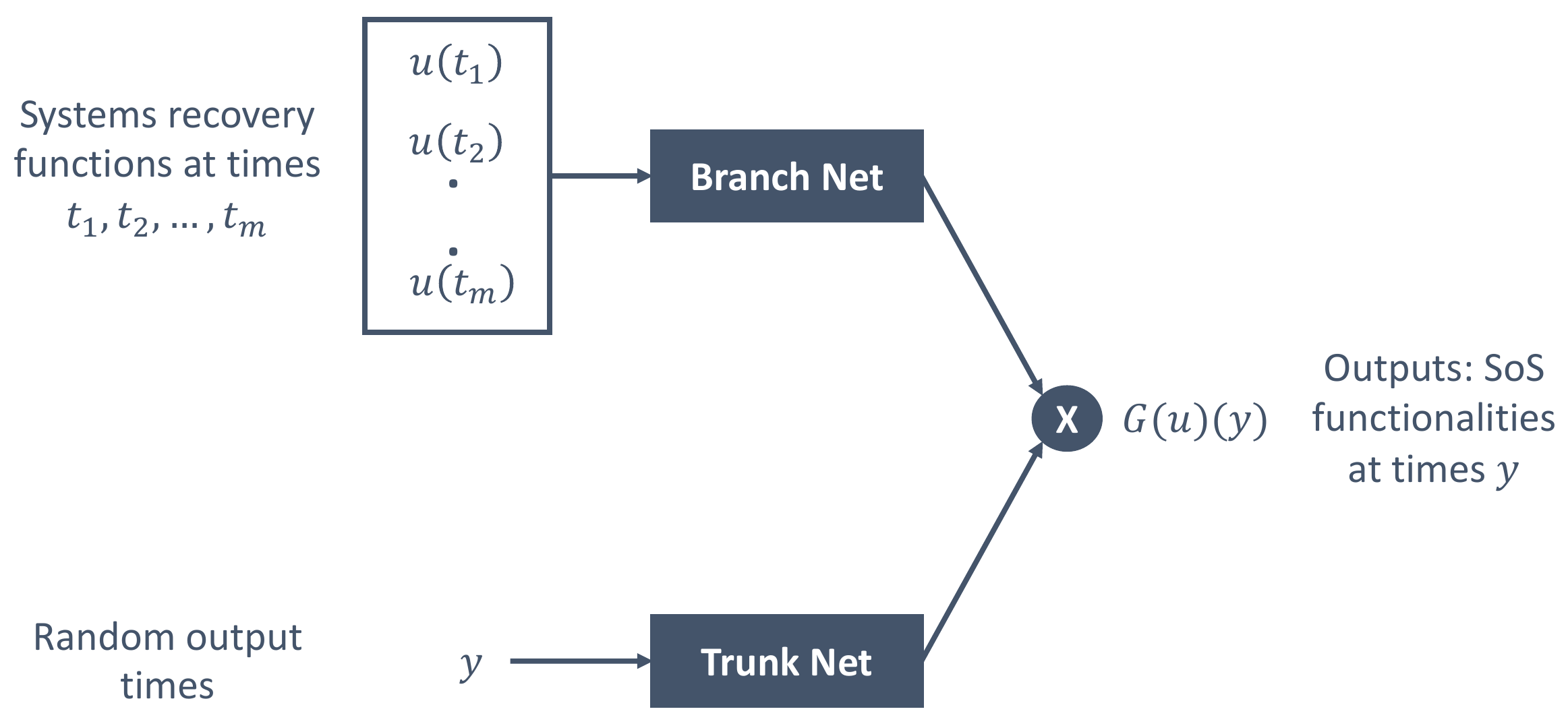}}
\caption{Architecture of DeepONets, inputs $\big(\{u(t_1),~u(t_2),\dots,u(t_m)\}$; $y\big)$, and outputs $\big(G(u)(y)\big)$. This figure is adapted from Lu et al. \cite{Lu2021a}.}
\label{fig:DeepONet}
\end{figure}

DeepONets identify nonlinear dynamic systems of the form in equation \eqref{eqn:DONet_1} using trunk and branch networks \cite{Lu2021a}. The trunk network takes as input $y$ and outputs $\{\mathcal{T}_1,~\mathcal{T}_2,\dots,\mathcal{T}_p\} \in \mathbb{R}^p$. There will be $p$ branch networks and each one of them takes as the input $\{u(t_1),~u(t_2),\dots,u(t_m)\}$ and outputs $b_k \in \mathbb{R}$. The prediction of DeepONets is given as:

\begin{equation}
    \label{eqn:DONet_2}
    G(u)(y) \approx \sum_{k=1}^p b_k~\mathcal{T}_k + b_0
\end{equation}

\noindent where, $b_0$ is a bias term. The requirement for the training data for DeepONets is that the sensor locations $\{t_1,~t_2,\dots,t_m\}$ are same for all the function values $u(t_m)$. The output locations $y$ can be random. Each output location is associated with the output $G(u)(y)$. Figure \ref{fig:DeepONet} presents the architecture of DeepONets, inputs, and outputs. 

For DeepONet training using SoS recovery data, the branch network takes as inputs individual systems recovery curves. These recovery curves depend on the external or internal hazard and its magnitude and can vary, as discussed in Section \ref{sec:results}. For a set of systems, all the functionality values have to be measured at the same times. Next, for the trunk network, the output times can be random or fixed to a set of values. Corresponding to these output times are the SoS functionality values. We used the DeepXDE \cite{Lu2021b} package for implementing DeepONets with systems and SOS recovery data.

\section{Results: Sixteen state system-of-systems}\label{sec:results}

We applied DeepONets to predict the recovery of a SoS with four systems. This example is obtained from section 8 of Dhulipala et al. \cite{Dhulipala2021a}. These four systems are generically defined. The SoS will have sixteen discrete functionality recovery states. The first state represents none of the four systems are functional. The next four states represent only one system is functional. The next six states represent two systems are functional. The next four states represent four systems are functional. The final state represents all the systems are functional. For the present demonstration, the interdependencies between the four systems are defined such that any system has the same impact on the SoS functionality when it is non-functional. This means, the $\pmb{F}$ vector in equation \eqref{eqn:SoS_Rec_2} has the sixteen components with the first component being 0, the next four components being 0.25, the next six components being 0.5, the next four components being 0.75, and the final component being 1.

\subsection{Identical recovery functions for systems}

\begin{figure}[htbp]
\centerline{\includegraphics[scale=0.5]{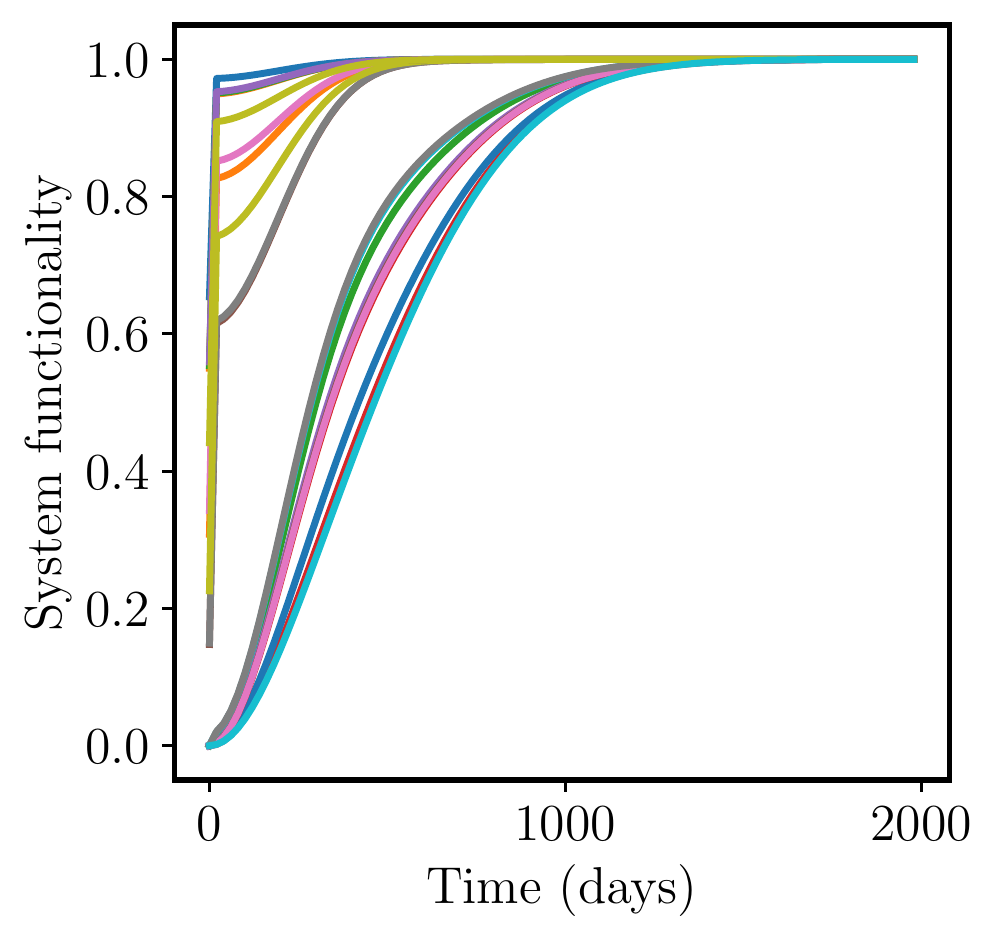}}
\caption{Twenty random recovery functions of the systems which are the inputs for simulating SoS recovery curves. These functions are part of the training data for DeepONets.}
\label{fig:Results_TrainInput}
\end{figure}

\begin{figure}[htbp]
\centerline{\includegraphics[scale=0.5]{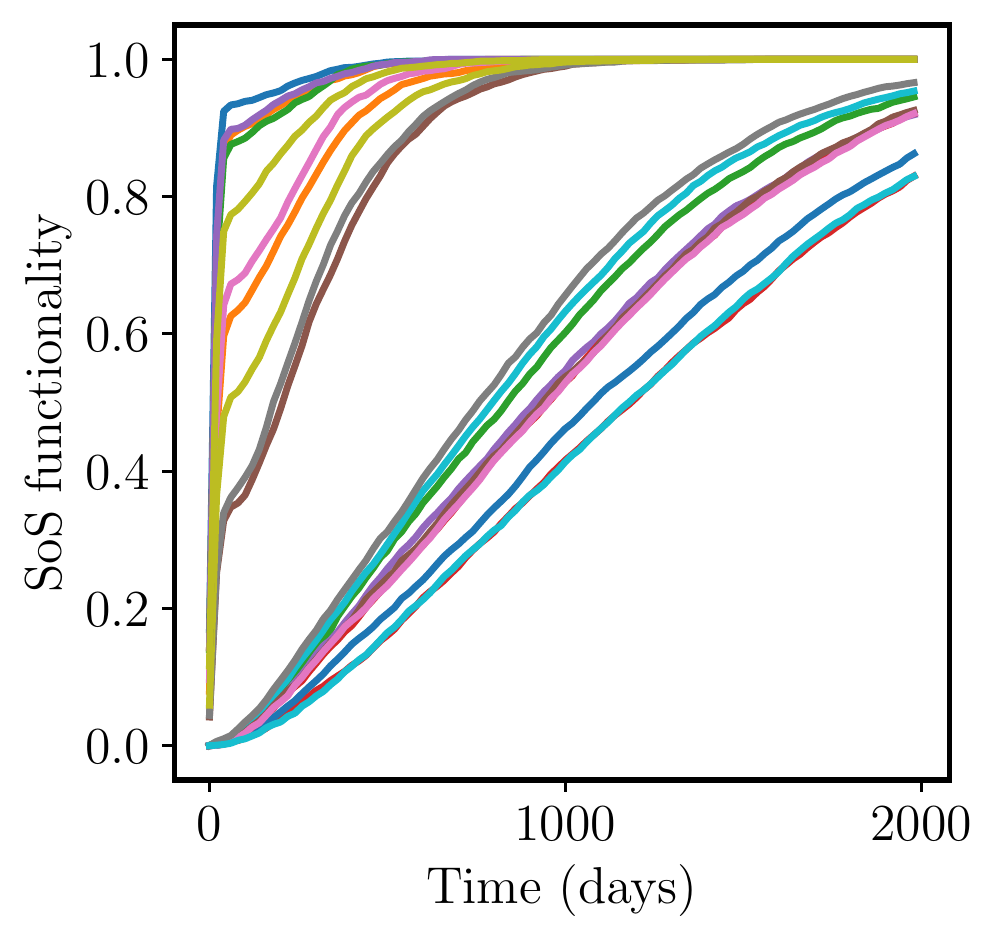}}
\caption{Twenty SoS recovery curves corresponding to the system recovery curves presented in Figure \ref{fig:Results_TrainInput}. These recovery curves are part of the training data for DeepONets.}
\label{fig:Results_TrainOutput}
\end{figure}

First, it is assumed that all the four systems have identical functionality recovery curves. We generated 20 random recovery functions for the systems and simulated the corresponding SoS recovery curves using a Monte Carlo method for solving equation \eqref{eqn:SoS_Rec_3}. Figures \ref{fig:Results_TrainInput} and \ref{fig:Results_TrainOutput} present the recovery functions of the systems and the SoS, respectively. This 20 set of recovery functions constitute the training data for DeepONets. We also generated 200 random recovery functions for the systems and simulated the corresponding SoS recovery curves as the testing data for DeepONets. 

\begin{figure}[htbp]
\centerline{\includegraphics[scale=0.32]{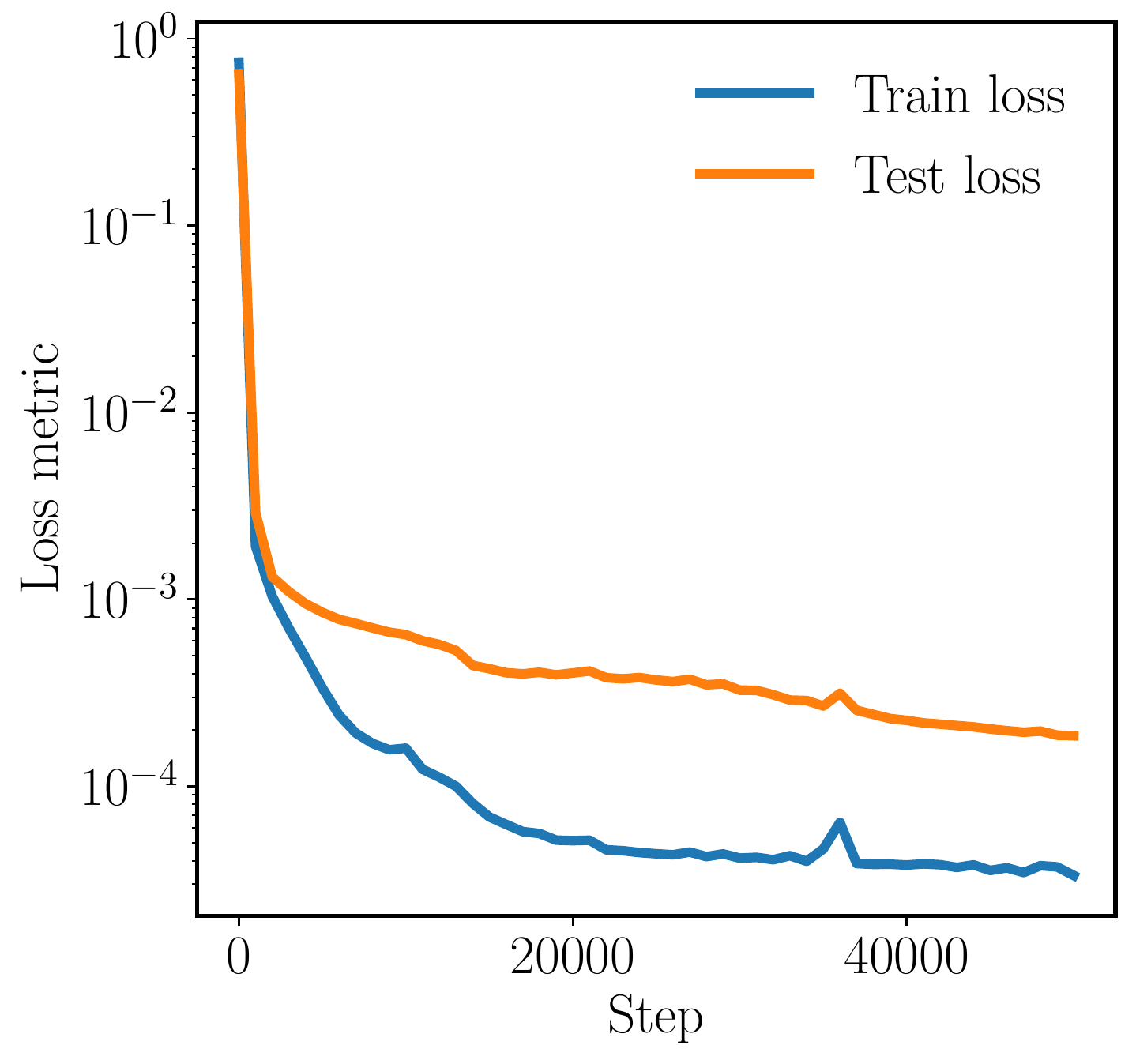}}
\caption{Loss metric evolution during the training of DeepONets.}
\label{fig:Results_LossHist}
\end{figure}

\begin{figure}[htbp]
\centerline{\includegraphics[scale=0.32]{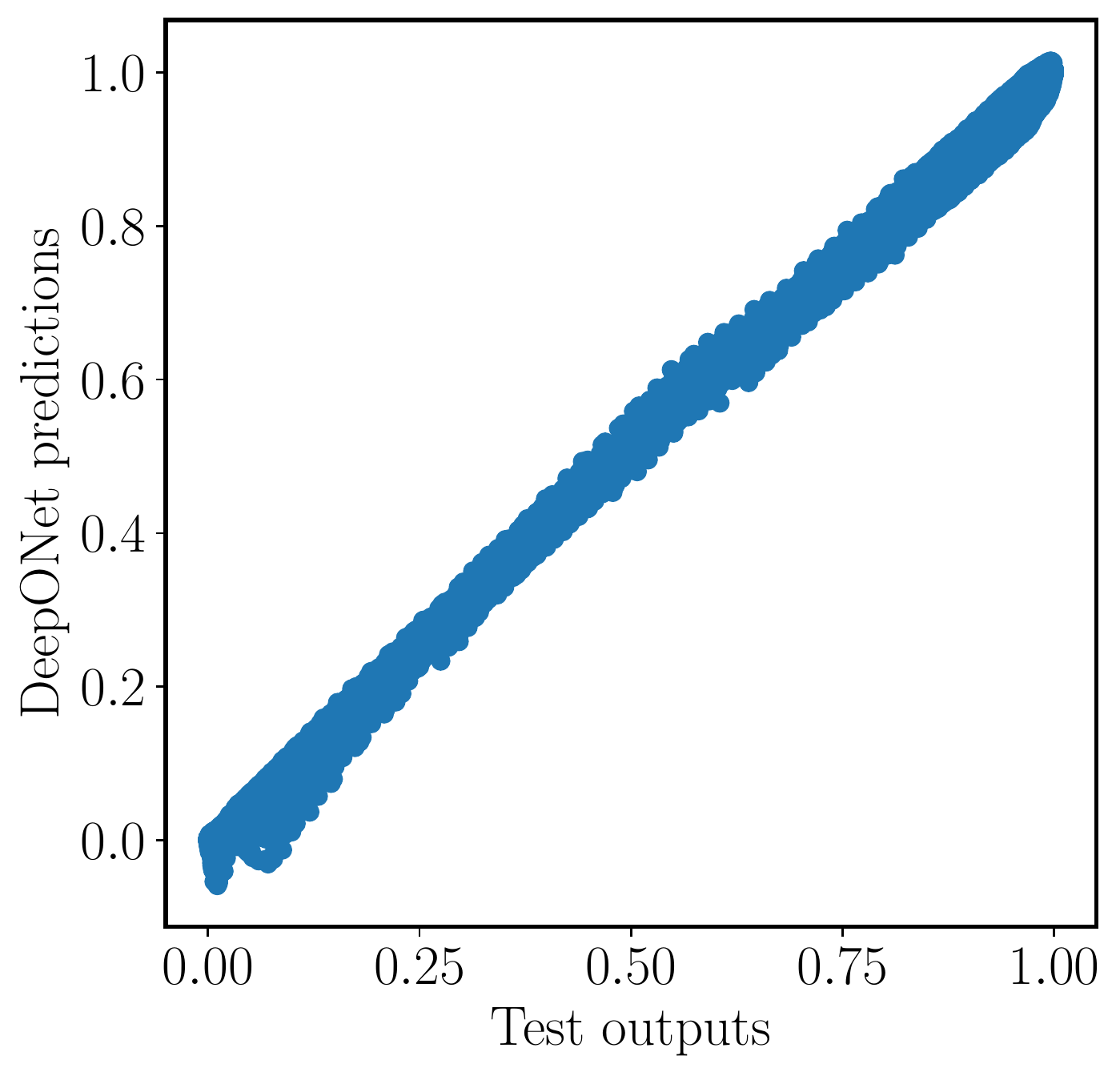}}
\caption{Scatter plot of the SoS functionality values from the test dataset and those predicted by DeepONets.}
\label{fig:Results_Scatter}
\end{figure}

\begin{figure}[htbp]
\centerline{\includegraphics[scale=0.32]{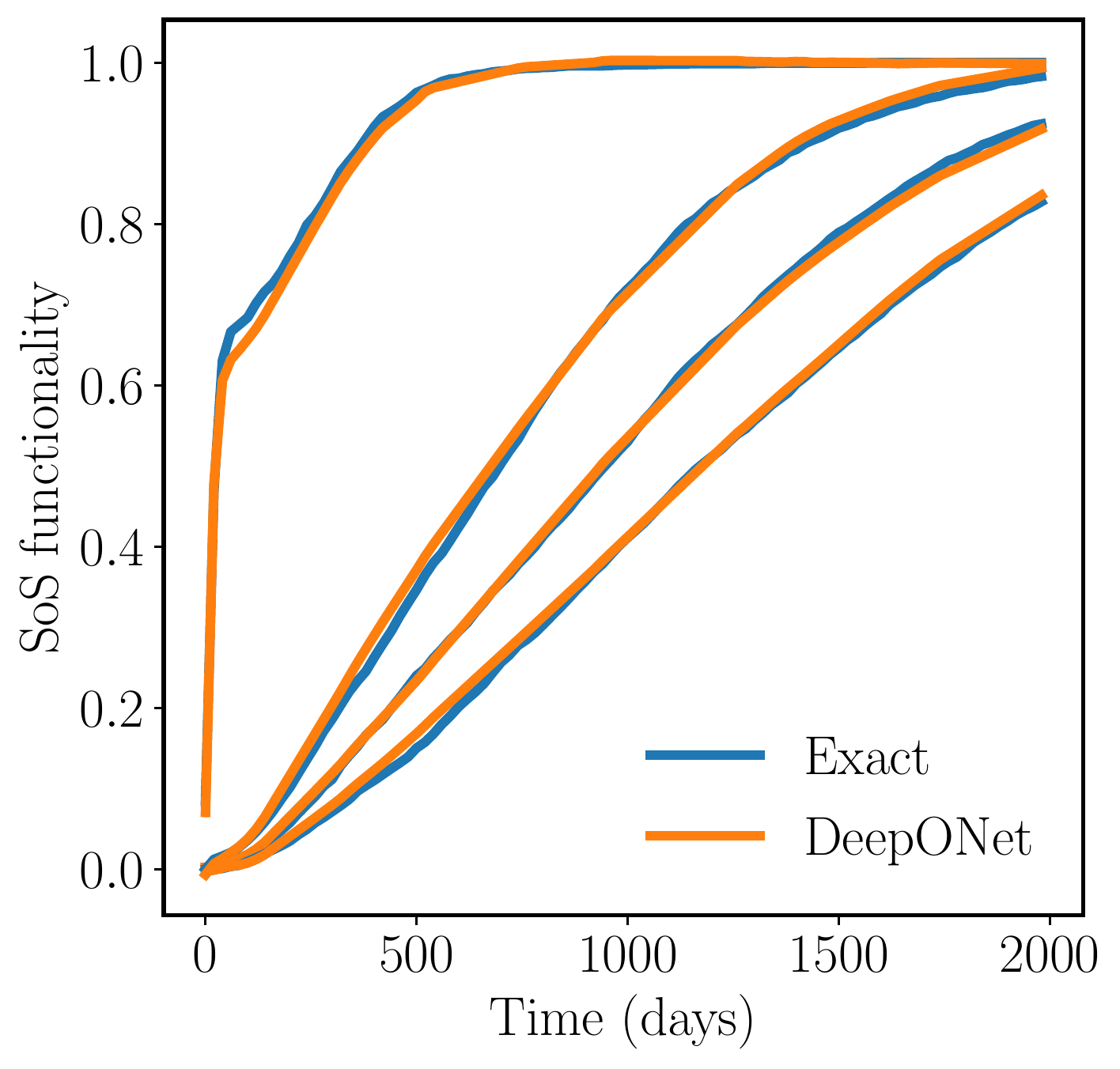}}
\caption{Comparison of the SoS recovery paths predicted by DeepONets with exact results for four data samples from the test dataset.}
\label{fig:Results_RecPaths}
\end{figure}

Figure \ref{fig:Results_LossHist} presents the loss metric evolution during the training of DeepONets. It is noticed that both the training and testing loss metrics are satisfactorily low at the end of training. Figure \ref{fig:Results_Scatter} presents a scatter plot of the SoS functionality values from the test dataset and those predicted by DeepONets. It is observed that the exact and predicted values have an excellent correlation given that the DeepONets are only trained on 20 training data samples. Figure \ref{fig:Results_RecPaths} presents a comparison of the SoS recovery paths predicted by DeepONets with exact results for four data samples from the test dataset. Overall, DeepONets seem to perform well for this example application of a SoS with four systems which have identical recovery functions.

\subsection{Disparate recovery functions for systems}

\begin{figure}[htbp]
\centerline{\includegraphics[scale=0.5]{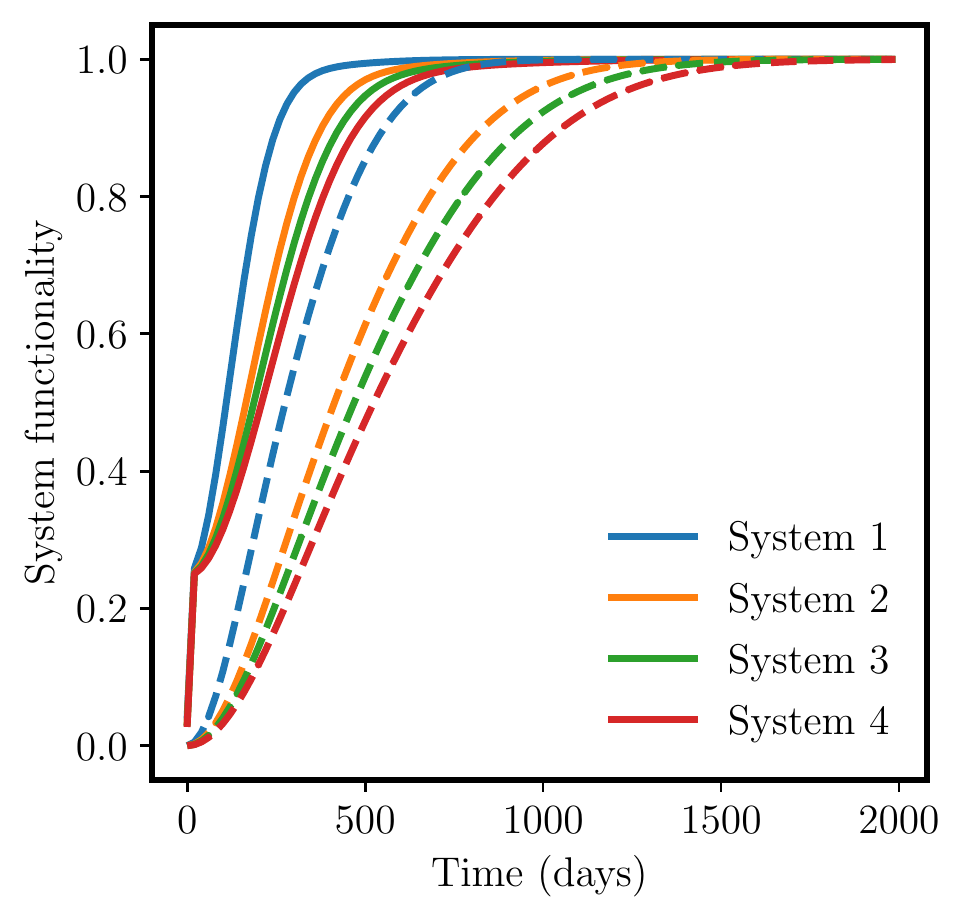}}
\caption{Two instances of random recovery function sets for the four systems. These functions are part of the training data for DeepONets. (solid: instance 1; dashed: instance 2)}
\label{fig:Results_TrainInput_Disp}
\end{figure}

\begin{figure}[htbp]
\centerline{\includegraphics[scale=0.5]{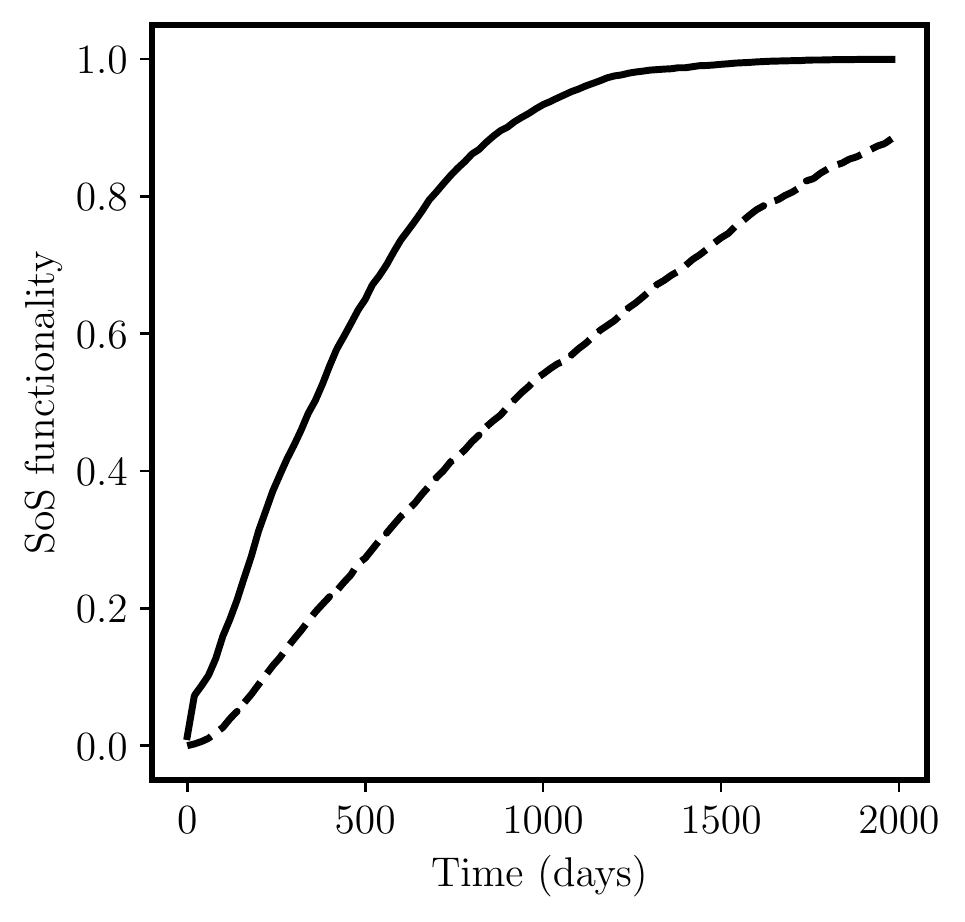}}
\caption{Two instances of SoS recovery curves corresponding to the system recovery curve sets presented in Figure \ref{fig:Results_TrainInput_Disp}. These recovery curves are part of the training data for DeepONets. (solid: instance 1; dashed: instance 2)}
\label{fig:Results_TrainOutput_Disp}
\end{figure}

We now consider the case when the four systems have disparate recovery functions. We generated 20 random recovery function sets for the systems and generated the corresponding SoS recovery curves using Monte Carlo which constitues the training data. Figure \ref{fig:Results_TrainInput_Disp} presents two instances of the system recovery function sets. Figure \ref{fig:Results_TrainOutput_Disp} presents the corresponding SoS recovery curves. We also generated 20 additional random recovery function sets for the systems and simulated the corresponding SoS recovery curves as the testing data for DeepONets. 

\begin{figure}[htbp]
\centerline{\includegraphics[scale=0.5]{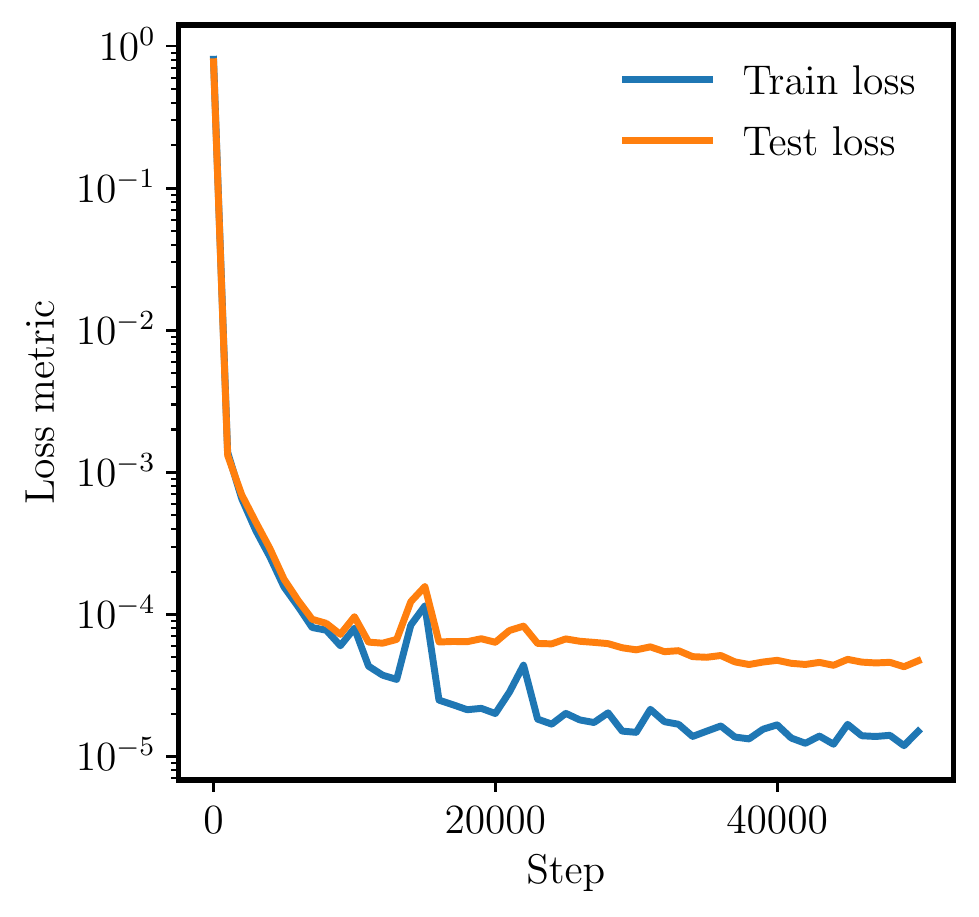}}
\caption{Loss metric evolution during the training of DeepONets. These results are for the case when the systems have disparate recovery functions.}
\label{fig:Results_LossHist_Disp}
\end{figure}

\begin{figure}[htbp]
\centerline{\includegraphics[scale=0.5]{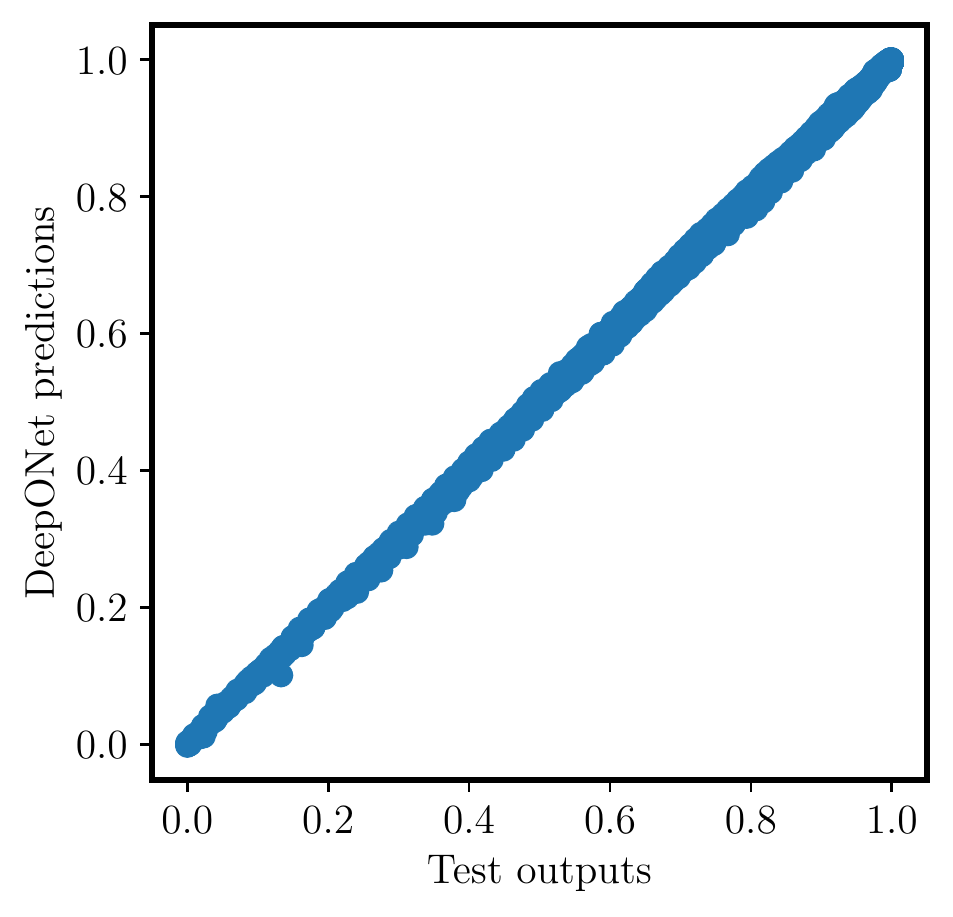}}
\caption{Scatter plot of the SoS functionality values from the test dataset and those predicted by DeepONets. These results are for the case when the systems have disparate recovery functions.}
\label{fig:Results_Scatter_Disp}
\end{figure}

\begin{figure}[htbp]
\centerline{\includegraphics[scale=0.5]{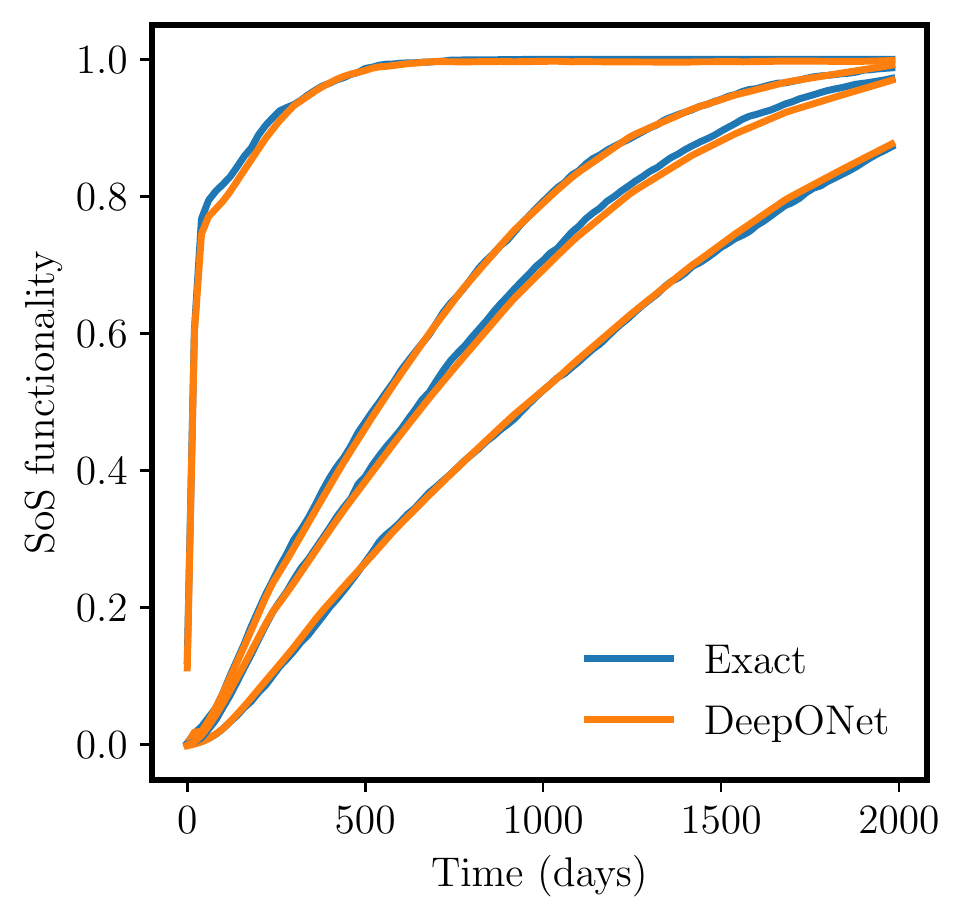}}
\caption{Comparison of the SoS recovery paths predicted by DeepONets with exact results for four data samples from the test dataset. These results are for the case when the systems have disparate recovery functions.}
\label{fig:Results_RecPaths_Disp}
\end{figure}

Figure \ref{fig:Results_LossHist_Disp} presents the loss metric evolution during the training of DeepONets. Figure \ref{fig:Results_Scatter_Disp} presents a scatter plot of the SoS functionality values from the test dataset and those predicted by DeepONets. It is observed that the exact and predicted values have an excellent correlation even for the case when the individual systems have disparate recovery functions. Figure \ref{fig:Results_RecPaths_Disp} presents a comparison of the SoS recovery paths predicted by DeepONets with exact results for four data samples from the test dataset. Overall, DeepONets seem to perform well for the case with disparate recovery functions for the systems. Future work includes expanding DeepONets application to more complex SoS with many systems.

\section{Conclusions and future work}\label{sec:conc}

In this study, we applied DeepONets to accelerate the recovery modeling of interdependent systems. When applied to an example problem of recovery modeling of a sixteen SoS, DeepONets perform satisfactorily despite not being trained on an exhaustive set of training data. Since the form of governing equations DeepONets identify from data and the governing equation of SoS recovery are similar, DeepONets performed well. Future work includes expanding the application of DeepONets to larger SoS with many interdependent systems.

\section*{Acknowledgments}

We thank Timothy McJunkin, Bjorn Vaagensmith, and Sam Yang for their valuable feedback on this work.

This research is supported through the INL Laboratory Directed Research \& Development (LDRD) Program under DOE Idaho Operations Office Contract DE-AC07-05ID14517.

\bibliography{references.bib}
\bibliographystyle{IEEEtran}

\end{document}